\documentclass[11pt,letterpaper]{article}
\usepackage{acl2016}
\usepackage{times}
\usepackage{latexsym}
\def\aclpaperid{622} 

\aclfinalcopy

\usepackage{times}
\usepackage{url}
\usepackage{latexsym}
\usepackage{enumitem}
\setlist{nosep} 
\usepackage{CJKutf8}
\usepackage{graphicx}
\usepackage{color}
\usepackage{comment}

\newcommand{\bleu}{\textsc{Bleu} }

\title{Transfer Learning for Low-Resource Neural Machine Translation}

\author{Barret Zoph$^{1}$, Deniz Yuret$^{2}$, Jonathan May$^{1}$, Kevin Knight$^{3}$ \\
$^{1}$Information Sciences Institute, University of Southern California \\
{\tt \{zoph, jonmay\}@isi.edu}\\
$^{2}$Computer Engineering,  Ko\c{c} University\\
{\tt dyuret@ku.edu.tr} \\
$^{3}$Information Sciences Institute \& \\
Computer Science Department, University of Southern California \\
{\tt knight@isi.edu}}

\date{}

\begin{document}
\maketitle

\begin{abstract}
The encoder-decoder framework for neural machine translation (NMT) has been shown effective in large data scenarios, but is much less effective for low-resource languages.  We present a transfer learning method that significantly improves \bleu scores across a range of low-resource languages.  Our key idea is to first train a high-resource language pair (the {\em parent model}), then transfer some of the learned parameters to the low-resource pair (the {\em child model}) to initialize and constrain training.  Using our transfer learning method we improve baseline NMT models by an average of 5.6 \bleu on four low-resource language pairs.  Ensembling and unknown word replacement add another 2 \bleu which brings the NMT performance on low-resource machine translation close to a strong syntax based machine translation (SBMT) system, exceeding its performance on one language pair. Additionally, using the transfer learning model for re-scoring, we can improve the SBMT system by an average of 1.3 \textsc{Bleu}, improving the state-of-the-art on low-resource machine translation.
\end{abstract}

\section{Introduction}

Neural machine translation (NMT) \cite{sutskever2014sequence} is a promising paradigm for extracting translation knowledge from parallel text.  NMT systems have  achieved competitive accuracy rates under large-data training conditions for language pairs such as English-French.  However, neural methods are data-hungry and learn poorly from low-count events.   This behavior makes vanilla NMT a poor choice for low-resource languages, where parallel data is scarce.  Table~\ref{low-d} shows that for 4~low-resource languages, a standard string-to-tree statistical MT system (SBMT) \cite{galley-EtAl:2004:HLTNAACL,galley-EtAl:2006:COLACL} strongly outperforms NMT, even when NMT uses the state-of-the-art local attention plus feed-input techniques from \newcite{luong2015effective}. 



\begin{table}
\begin{tabular}{|l|r|r|r|r|}
\hline Language & Train & Test & SBMT & NMT \\
& size & size & \bleu & \bleu \\ \hline 
Hausa & 1.0m & 11.3K & 23.7 & 16.8  \\ \hline
Turkish & 1.4m & 11.6K & 20.4 & 11.4  \\ \hline
Uzbek & 1.8m & 11.5K & 17.9 & 10.7 \\ \hline
Urdu & 0.2m & 11.4K & 17.9 & 5.2 \\ \hline
\end{tabular}
\caption{NMT models with attention are outperformed by standard string-to-tree statistical MT (SBMT) when translating low-resource languages into English.  Train/test bitext corpus sizes are given in word tokens on the English side.  Single-reference, case-insensitive \bleu scores are given for held-out test corpora.}
\label{low-d}
\end{table}

In this paper, we give a method for substantially improving NMT results on these languages. Neural models  learn representations of their input that are often useful across tasks.  Our key idea is to first train a high-resource language pair, then use the resulting trained network (the {\em parent model}) to initialize and constrain training for our low-resource language pair (the {\em child model}).  We find that we can optimize our results by fixing certain parameters of the parent model, letting the rest be fine-tuned by the child model.  We report NMT improvements from transfer learning of 5.6 \bleu on average, and we provide an analysis of why the method works. The final NMT system approaches strong SBMT baselines in in all four language pairs, and exceeds  SBMT performance in one of them. Furthermore, we show that NMT is an exceptional re-scorer of `traditional' MT output; even NMT that on its own is worse than SBMT is consistently able to improve upon SBMT system output when incorporated as a re-scoring model.

We start by a brief description of our NMT model in Section~\ref{nmt}.  Section~\ref{tl} gives some background on transfer learning and explains how we use it to improve machine translation performance.  Our main experiments translating Hausa, Turkish, Uzbek and Urdu into English with the help of a French-English parent model are presented in Section~\ref{experiments}.  Section~\ref{analysis} explores alternatives to our model to enhance understanding.  We find that the choice of parent language pair affects performance, and provide an empirical upper bound on transfer performance using an artificial language.  We experiment with English-only language models, copy models, and word-sorting models to show that what we transfer goes beyond monolingual information and using a translation model trained on bilingual corpora as a parent is essential. We show the effects of freezing, fine-tuning, and smarter initialization of different components of the attention-based NMT system during transfer.  We compare the learning curves of transfer and no-transfer models showing that transfer solves an overfitting problem, not a search problem.  We summarize our contributions in Section~\ref{conclusion}.

\section{NMT Background}
\label{nmt} 
\begin{figure}[t]
\begin{center}
\includegraphics[width=8cm]{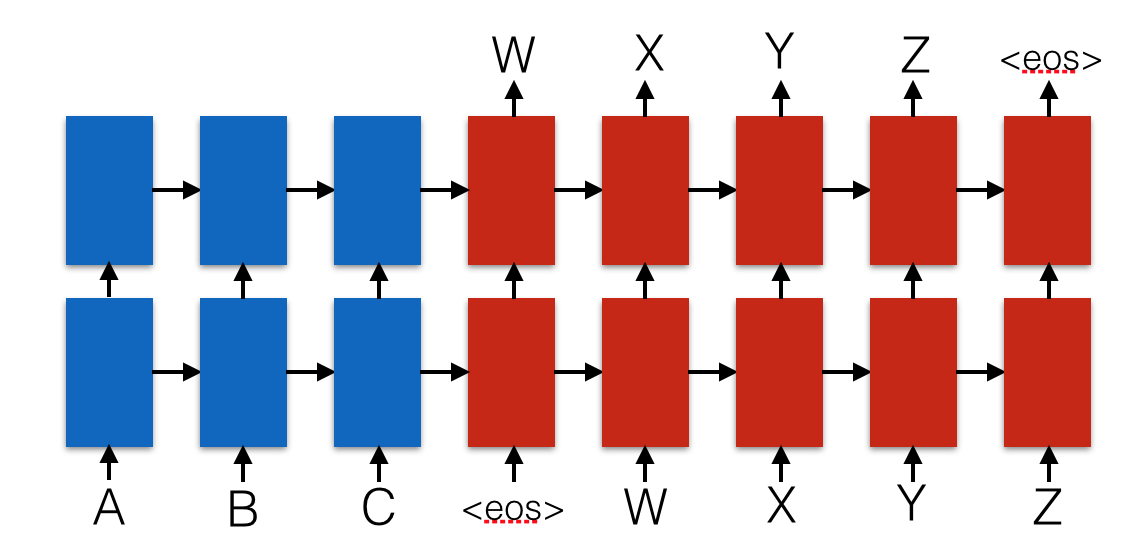}
\end{center}
\vspace{-0.1in}
\caption[]{The encoder-decoder framework for neural machine translation (NMT) \cite{sutskever2014sequence}.  Here, a source sentence C B A (presented in reverse order as A B C) is translated into a target sentence W X Y Z.  At each step, an evolving real-valued vector summarizes the state of the encoder (blue) and decoder (red).  Not shown here are the attention connections present in our model used by the decoder to access encoder states.}
\label{enc-dec}
\end{figure}

In the neural encoder-decoder framework for MT \cite{neco1997asynchronous,castano1997connectionist,sutskever2014sequence,bahdanau2014neural,luong2015effective}, we use a recurrent neural network (encoder) to convert a source sentence into a dense, fixed-length vector.  We then use another recurrent network (decoder) to convert that vector to a target sentence. In this paper, we use a two-layer encoder-decoder system (Figure~\ref{enc-dec}) with long short-term memory (LSTM) units \cite{lstm} trained to optimize maximum likelihood (via a softmax layer) with back-propagation through time \cite{btt}. 
Additionally, we use an attention mechanism that allows the target decoder to look back at the source encoder, specifically the local attention plus feed-input model from \newcite{luong2015effective}.

\section{Transfer Learning}
\label{tl}

Transfer learning uses knowledge from a learned task to improve the performance on a related task, typically reducing the amount of required training data \cite{torrey2009transfer,pan2010survey}.  In natural language processing, transfer learning methods have been successfully applied to speech recognition, document classification and sentiment analysis \cite{wang2015transfer}.  Deep learning models discover multiple levels of representation, some of which may be useful across tasks, which makes them particularly suited to transfer learning \cite{bengio2012deep}.  For example, \newcite{cirecsan2012transfer} use a convolutional neural network to recognize handwritten characters and show positive effects of transfer between models for Latin  and Chinese characters.  Ours is the first study to apply transfer learning to neural machine translation.


The transfer learning approach we use is simple and effective. We first train an NMT model on a dataset where there is a large amount of bilingual data (e.g., French-English), which we call the {\em parent model}. Next, we initialize an NMT model with the already-trained parent model. This new model is then trained on a dataset with very little bilingual data (e.g., Uzbek-English), which we call the {\em child model}.  This means that the low-data NMT model will not start with random weights, but with the weights from the parent model.

A justification for this approach is that in scenarios where we have limited training data, we need a strong prior distribution over models. The parent model trained on a large amount of bilingual data can be considered an anchor point, the peak of our prior distribution in model space. When we train the child model initialized with the parent model, we fix parameters likely to be useful across tasks so that they will not be changed during child-model training. In the French-English to Uzbek-English example, as a result of the initialization, the English word embeddings from the parent model are copied, but the Uzbek words are initially mapped to random French embeddings. The English embeddings should be kept but the Uzbek embeddings should be modified during training of the child model.  Freezing certain portions of the parent model and fine tuning others can be considered a hard approximation to a tight prior or strong regularization applied to some of the parameters. We also experiment with ordinary L2 regularization, but find it does not significantly improve over the parameter freezing described above. 

Our method results in large \bleu increases for a variety of low resource languages. In one of the four language pairs our NMT system using transfer beats a strong SBMT baseline. Not only do these transfer models do well on their own, they also give large gains when used for rescoring $n$-best lists ($n = 1000$) from the SBMT system.  Section~\ref{experiments} details these results.

\section{Experiments}
\label{experiments}
To evaluate how well our transfer method works we apply it to a variety of low-resource languages, both stand-alone and for re-scoring a strong SBMT baseline. We report large \bleu increases across the board with our transfer method.

For all of our experiments with low-resource languages we use French as the parent source language and for child source languages we use Hausa, Turkish, Uzbek, and Urdu. The target language is always English.  Table~\ref{low-d} shows parallel training data set sizes for the child languages, where the language with the most data has only 1.8m English tokens. For comparison, our parent French-English model uses a training set with 300 million English tokens and achieves 26 \bleu on the development set. Table~\ref{low-d} also shows the SBMT system scores along with the NMT baselines that do not use transfer. There is a large gap between the SBMT and NMT systems without using our transfer method. 

The SBMT system used in this paper is a string-to-tree statistical machine translation system \cite{galley-EtAl:2006:COLACL,galley-EtAl:2004:HLTNAACL}. In this system there are two count-based 5-gram language models. One is trained on the English side of the WMT 2015 English-French dataset and the other is trained on the English side of the low-resource bitext. Additionally, the SBMT models use thousands of sparsely-occurring, lexicalized syntactic features \cite{new-features}.

For our NMT system, we use development sets for Hausa, Turkish, Uzbek, and Urdu to tune the learning rate, parameter initialization range, dropout rate and hidden state size for all the experiments. For training we use a minibatch size of 128, hidden state size of 1000, a target vocabulary size of 15K and a source vocabulary size of 30K. The child models are trained with a dropout probability of 0.5 as in \newcite{dropout}. The common parent model is trained with a dropout probability of 0.2. The learning rate used for both child and parents is 0.5 with a decay rate of 0.9 when the development perplexity does not improve.  The child models are all trained for 100 epochs. We re-scale the gradient when the gradient norm is greater than~5. The initial parameter range is \mbox{$[$-0.08,~+0.08$]$}.

\subsection{Transfer Results}

\begin{table}
\begin{tabular}{|l|r|r|r|r|}
\hline Language & SBMT & NMT & Xfer & Final\\ \hline
Hausa & 23.7 & 16.8 & 21.3 & 24.0 \\ \hline
Turkish & 20.4 & 11.4 & 17.0 & 18.7 \\ \hline
Uzbek & 17.9 & 10.7 & 14.4 & 16.8 \\ \hline
Urdu & 17.9 & 5.2 & 13.8 & 14.5 \\ \hline
\end{tabular}
\caption{Our method significantly improves NMT results for the translation of low-resource languages into English.  Results show test-set \bleu scores. The `NMT' column shows results without transfer, and the `Xfer' column shows results with transfer. The `Final' column shows \bleu after we ensemble 8 models and use unknown word replacement.}
\label{adaptation-results}
\end{table}

The results for our transfer learning method applied to the four languages above are in Table~\ref{adaptation-results}. The parent models were trained on the WMT 2015 \cite{bojar-EtAl:2015:WMT} French-English corpus for 5 epochs. Our baseline NMT systems (`NMT' column) all receive a large \bleu improvement when using the transfer method (the `Xfer' column) with an average \bleu improvement of 5.6.  Additionally, when we use unknown word replacement from \newcite{luong-EtAl:2015:ACL-IJCNLP} and ensemble together 8 models (the `Final' column) we further improve upon our \bleu scores, bringing the average \bleu improvement to 7.5. Overall our method allows the NMT system to reach competitive scores and beat the SBMT system in one of the four language pairs.

\subsection{Re-scoring Results}

\begin{table}
\begin{tabular}{|l|r|r|r|r|}
\hline Setting & SBMT & NMT & Xfer & LM \\ \hline
Hausa & 23.7 & 24.5 & {\bf 24.8} & 23.6 \\ \hline
Turkish & 20.4 & 21.4 & {\bf 21.8} & 21.1 \\ \hline
Uzbek & 17.9 & 19.5 & {\bf 19.5} & 17.9 \\ \hline
Urdu & 17.9 & 18.2 & {\bf 19.1} & 18.2 \\ \hline
\end{tabular}
\caption{Our transfer method applied to re-scoring output $n$-best lists from the SBMT system. Additionally, the `LM' column shows the results when an RNN LM was trained on the large English corpus and used to re-score the $n$-best list.}
\label{re-score}
\end{table}

We also use the NMT model with transfer learning to re-score output $n$-best lists ($n = 1000$) from the SBMT system.  Table~\ref{re-score} shows the results of re-scoring. Transfer NMT models give the highest gains over using re-scoring with a neural language model or an NMT system that does not use transfer.  The neural language model is an LSTM RNN with 2 layers and 1000 hidden states. It has a target vocabulary of 100K and is trained using noise-contrastive estimation \cite{mnih2012fast,vaswani2013decoding,baltescu2014pragmatic,williams15}. Additionally, it is trained using dropout with a dropout probability of 0.2 as in \cite{dropout}. From re-scoring with the transfer model, we get an improvement of 1.1--1.6 \bleu points above the strong SBMT system.

We ran a number of additional experiments to understand what components of our final transfer model significantly contribute to the overall result.  Section~\ref{analysis} details these experiments.

\section{Analysis}
\label{analysis}

We analyze the effects of using different parent models, regularizing different parts of the child model and trying different regularization techniques.

\subsection{Different Parent Languages}

\begin{table}
\begin{tabular}{|l|r|r|r|r|}
\hline Language pair & Role & Train & Dev & Test \\ 
  & & Size  & Size & Size \\ \hline 
Spanish-English & child & 2.5m & 58k & 59k \\ \hline
French-English & parent & 53m & 58k & 59k \\ \hline
German-English & parent & 53m & 58k & 59k \\ \hline
\end{tabular}
\caption{Data used for a low-resource Spanish-English task. Sizes are numbers of word tokens on the English side of the bitext.
}
\label{spa-eng-data}
\end{table}

In the above experiments we use French-English as the parent language pair.  Here, we experiment with different parent languages. In this set of experiments we use Spanish-English as the child language pair, rather then Hausa, Turkish, Uzbek, or Urdu. A description of the data used in this section is presented in Table~\ref{spa-eng-data}.

Our experimental results are shown in Table~\ref{different-languages}, where we use French and German as parent languages.  If we just train a model with no transfer on a small Spanish-English training set we get a \bleu score of 16.4. When using our transfer method using French and German as parent languages, we get \bleu scores of 31.0 and 29.8 respectively. As expected, French is a better parent than German for Spanish, which could be the result of the parent language being more similar to the child language.

Overall, we can see that the choice of parent language can make a difference in the \bleu score, so in our Hausa, Turkish, Uzbek, and Urdu experiments, a parent language more optimal than French might improve results. 

\begin{table}
\begin{tabular}{|l|r|r|}
\hline Parent & \bleu & PPL\\ \hline
none & 16.4 & 15.9\\ \hline
French-English & {\bf 31.0} & {\bf 5.8} \\ \hline
German-English & 29.8 & 6.2 \\ \hline
\end{tabular}
\caption{For a low-resource Spanish-English task, we experiment with several choices of parent model: none, French-English, and German-English.  We hypothesize that French-English is best because French and Spanish are similar.}
\label{different-languages}
\end{table}

\subsection{Effects of having Similar Parent Language}

\begin{table}
\begin{tabular}{|l|r|r|r|}
\hline Model & Train & \bleu & PPL \\ 
& size & & \\ \hline
Uzbek-English & 1.8m & 10.7 & 22.4 \\ \hline
Uzbek-English transfer & 1.8m & {\bf 15.0 (+4.3)} & {\bf 13.9} \\ \hline \hline
French'-English & 1.8m & 13.3 & 28.2 \\ \hline
French'-English transfer & 1.8m & {\bf 20.0 (+6.7)} & {\bf 10.9} \\ \hline 
\end{tabular}
\caption{A better match between parent and child languages should improve transfer results. We devised a child language called French', identical to French except for word spellings.  We observe that French transfer learning helps French' (13.3$\rightarrow$20.0) more than it helps Uzbek (10.7$\rightarrow$15.0).}
\label{french-prime}
\end{table}

Next, we look at a best-case scenario in which the parent language is as similar as possible to the child language. 
Here we devise a synthetic child language (called French') which is exactly like French, except its vocabulary is shuffled randomly. (e.g., ``internationale'' is now ``pomme'', etc). This language, which looks unintelligible to human eyes, nevertheless has the same distributional and relational properties as actual French, i.e. the word that, prior to vocabulary reassignment, was `roi' (king) is likely to share distributional characteristics, and hence embedding similarity, to the word that, prior to reassignment, was `reine' (queen). Such a language should be the ideal parent model.

The results of this experiment are shown in Table~\ref{french-prime}. We get a 4.3 \bleu improvement with an unrelated parent (i.e. French-parent and Uzbek-child), but we get a 6.7 \bleu improvement with a `closely related' parent (i.e. French-parent and French'-child). We conclude that the choice of parent model can have a strong impact on transfer models, and choosing better parents for our low-resource languages (if data for such parents can be obtained) could improve the final results.

\subsection{Ablation Analysis}

\begin{figure}
\includegraphics[width=\linewidth]{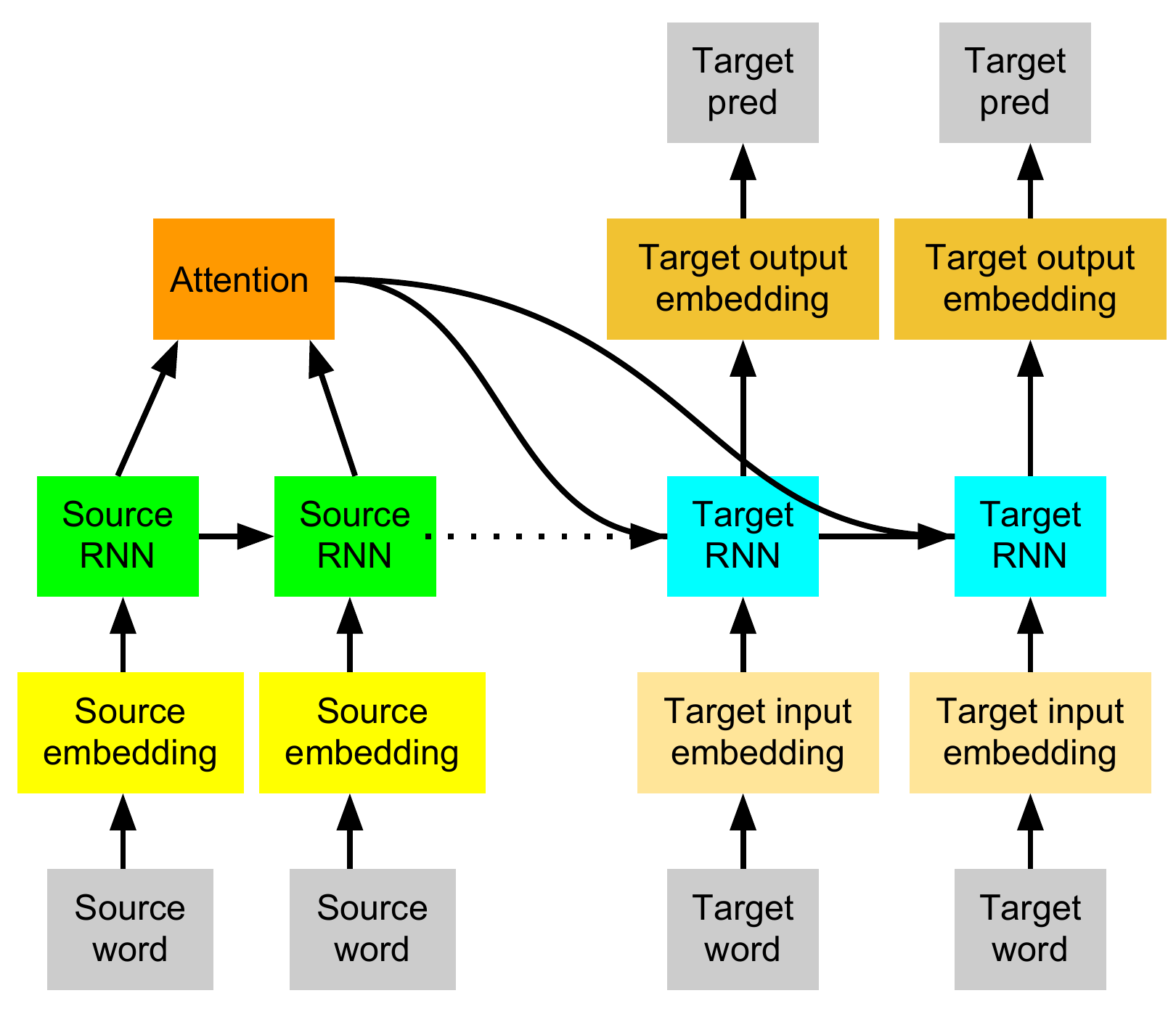}
\caption{Our NMT model architecture, showing six blocks of parameters, in addition to source/target words and predictions. During transfer learning, we expect the source-language related blocks to change more than the target-language related blocks.}
\label{rnncomponents}
\end{figure}

\begin{table}
\begin{tabular}{|l|r|r|}
\hline Setting & Dev  & Dev  \\ 
 &  \bleu &  PPL \\ \hline
No retraining & 0.0 & 112.6 \\ \hline
Retrain source embeddings & 7.7 & 24.7 \\ \hline
+ source RNN & 11.8 & 17.0 \\ \hline
+ target RNN & 14.2 & 14.5 \\ \hline
+ target attention & {\bf 15.0} & 13.9 \\ \hline
+ target input embeddings & 14.7 & {\bf 13.8} \\ \hline
+ target output embeddings & 13.7 & 14.4 \\ \hline
\end{tabular}
\caption{Starting with the parent French-English model (\bleu=24.4, PPL=6.2), we randomly assign Uzbek word types to French word embeddings, freeze various parameters of the neural network model, and allow Uzbek-English (child model) training to modify other parts.  The table shows how Uzbek-English \bleu and perplexity vary as we allow more parameters to be re-trained.}
\label{training-settings}
\end{table}

In all the above experiments, only the target input and output embeddings are fixed during training. In this section we analyze what happens when different parts of the model are fixed, in order to see what yields optimal performance. Figure~\ref{rnncomponents} shows a diagram of the components of a sequence-to-sequence model. Table~\ref{training-settings} shows how we begin to allow more components of the child NMT model to be trained and see the effect on performance in the model. We see that the optimal setting for transferring from French-English to Uzbek-English in terms of \bleu performance is to allow all of the components of the child model to be trained except for the input and output target embeddings.  

Even though we use this setting for our main experiments, the optimum setting is likely to be language- and corpus-dependent.  For Turkish, experiments show that freezing target attention parameters as well gives slightly better results.  For parent-child models with closely related languages we expect freezing, or strongly regularizing, more components of the model to give better results.



\subsection{Learning Curve}

\begin{figure}
\hspace*{-5mm}
\includegraphics[width=1.1\linewidth]{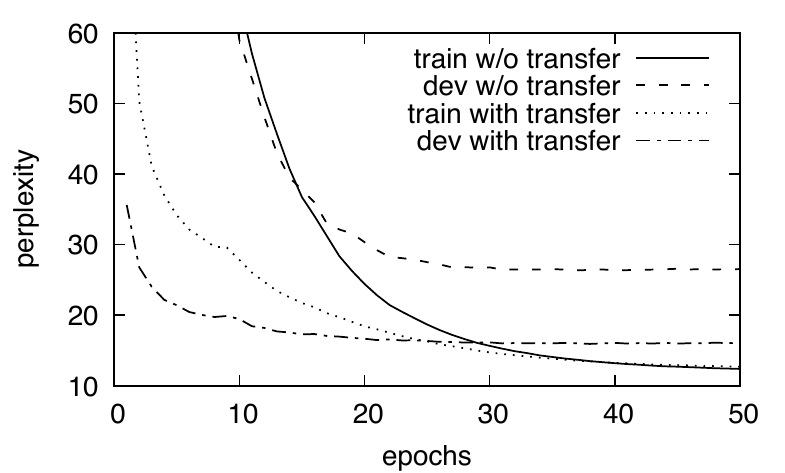}
\caption{Uzbek-English learning curves for the NMT attention model with and without transfer learning.  The training perplexity converges to a similar value in both cases.  However, the development perplexity for the transfer model is significantly better.}
\label{attvspre}
\end{figure}

In Figure~\ref{attvspre} we plot learning curves for both a transfer and a non-transfer model on training and development sets. We see that the final training set perplexities for both the transfer and non-transfer model are very similar, but the development set perplexity for the transfer model is much better. 

The fact that the two models start from and converge to very different points, yet have similar training set performances, indicates that our architecture and training algorithm are able to reach a good minimum of the training objective regardless of the initialization.  However, the training objective seems to have a large basin of models with similar performance and not all of them generalize well to the development set. The transfer model, starting with and staying close to a point known to perform well on a related task, is guided to a final point in the weight space that generalizes to the development set much better.

\subsection{Dictionary Initialization}

\begin{figure}
\hspace*{-5mm}
\includegraphics[width=1.1\linewidth]{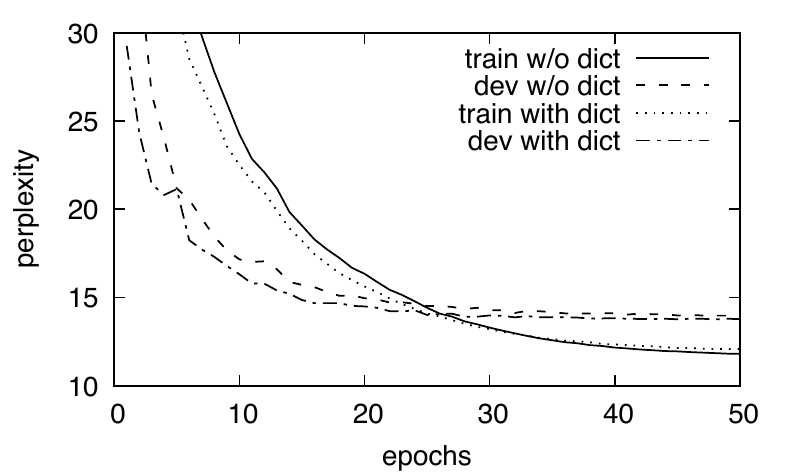}
\caption{Uzbek-English learning curves for the transfer model with and without dictionary-based assignment of Uzbek word types to French word embeddings (from the parent model).  Dictionary-based assignment enables faster improvement in early epochs, reaching 25 dev perplexity at epoch 2 vs 4.  However the final perplexities are similar, showing that the model is able to untangle the initial random Uzbek/French word-type mapping without help.}
\label{dictinit}
\end{figure}

Using the transfer method, we always initialize input language embeddings for the child model with randomly-assigned embeddings from the parent (which has a different input language). A smarter method might be to initialize child embeddings with similar parent embeddings, where similarity is measured by word-to-word t-table probabilities. To get these probabilities we compose Uzbek-English and English-French t-tables  obtained from the Berkeley Aligner \cite{liang06alignment}. We see from Figure~\ref{dictinit} that this dictionary-based assignment results in faster improvement in the early part of the training.  However the final performance is similar to our standard model, indicating that the training is able to untangle the dictionary permutation introduced by randomly-assigned embeddings.

\subsection{Different Parent Models}

\begin{table}
\begin{tabular}{|l|r|r|}
\hline Setting & \bleu & PPL \\ \hline
NMT & 10.7 & 22.4 \\ \hline
French-English transfer & {\bf 14.4} & {\bf 14.3} \\ \hline
English-English transfer & 5.3 & 55.8 \\ \hline
EngPerm-English transfer & 10.8 & 20.4 \\ \hline
LM transfer & 12.9 & 16.3 \\ \hline
\end{tabular}
\caption{ Transfer with parent models trained only on English data. The child data is the Uzbek-English corpus from Table~\ref{re-score}. The English-English parent learns to copy English sentences, and the EngPerm-English learns to un-permute scrambled English sentences. The  LM is a 2-layer LSTM RNN language model trained on the English corpus.}
\label{copy-perm-parents}
\end{table}

In the above experiments, we use a parent model trained on a large French/English bilingual corpus.  One might hypothesize that our gains come from exploiting the English half of the corpus as an additional language model resource.  Therefore, we explore transfer learning for the child model with parent models that only use the English side of the bilingual corpus. Table~\ref{copy-perm-parents} shows the results for these experiments where we train one parent model to copy English sentences (English-English) and another parent model to un-permute scrambled English sentences (EngPerm-English). Additionally, we train a parent model that is just an RNN language model. These results show that our transfer learning is not simply importing an English language model, but making use of translation parameters learned from the parent's large bilingual text.

\section{Conclusion}
\label{conclusion} 

Overall our transfer method improves NMT scores on low-resource languages by a large margin and allows our transfer NMT system to come close to the performance of a very strong SBMT system, even exceeding its performance on Hausa-English. In addition, we consistently and significantly improve state-of-the-art SBMT systems on low-resource languages when the transfer NMT system is used for re-scoring.  Our experiments suggest that there is still room for improvement in selecting parent languages that are more similar to child languages, provided data for such parents can be found.

\section{Acknowledgments}

This work was carried out with funding from DARPA (HR0011-15-C-0115) and
ARL/ARO (W911NF-10-1-0533).




\bibliographystyle{acl2016}
\bibliography{low-res,extra}
 
\end{document}